\documentclass{OAGM}

\usepackage{fullpage}
\usepackage[utf8x]{inputenc}
\usepackage[english]{babel}
\usepackage{amsmath}
\usepackage{amssymb}
\usepackage{amsthm}
\usepackage{latexsym}
\usepackage{psfrag}
\usepackage{graphicx}
\usepackage[font=small,labelfont=bf]{caption}
\usepackage{nameref}
\usepackage{enumerate}
\usepackage[]{algorithm2e}
\usepackage{cite}
\usepackage{hyperref}
\usepackage{tabularx}
\usepackage{wrapfig}
\usepackage{tikz}
\usetikzlibrary{spy}
\newcolumntype{C}{>{\centering\arraybackslash}X}
\OAGMarXiv{1505.01065}

\newcommand{\eqnref}[1]{(\ref{#1})}

\title{Smooth and iteratively Restore: A simple and fast edge-preserving smoothing model}
\author{Philipp Kniefacz\\Vienna University of Technology, Austria\and Walter Kropatsch\\Vienna University of Technology, Austria}

\begin{document}

\maketitle

\begin{abstract}
	In image processing, it can be a useful pre-processing step to smooth away small structures, such as noise or unimportant details, while retaining the overall structure of the image by keeping edges, which separate objects, sharp. Typically this edge-preserving smoothing process is achieved using edge-aware filters. However such filters may preserve unwanted small structures as well if they contain edges. In this work we present a novel framework for edge-preserving smoothing which separates the process into two different steps: First the image is smoothed using a blurring filter and in the second step the important edges are restored using a guided edge-aware filter. The presented method proves to deliver very good results, compared to state-of-the-art edge-preserving smoothing filters, especially at removing unwanted small structures. Furthermore it is very versatile and can easily be adapted to different fields of applications while at the same time being very fast to compute and therefore well-suited for real time applications.
\end{abstract}

\section{Introduction}
Edge-preserving smoothing is an approach to smooth across boundaries between regions of similar appearance while retaining strong edges. It is a technique to remove noise, weak edges and small details whereas the overall structure of the image should not be lost. Recent edge-preserving smoothing techniques, such as the joint bilateral filter \cite{petschnigg2004digital}, the guided filter \cite{guidedfilter} and the rolling guidance filter \cite{rollingguidance}, use a so called guidance image to perform this operation. \\
In this work we propose a novel framework for edge-preserving smoothing. Like the rolling guidance filter proposed in \cite{rollingguidance} it aims not only for edge-aware filtering, but also for scale-aware filtering. Our model splits the tasks of smoothing and edge preservation by using two different filters: a smoothing filter and a guided edge-aware filter. In the first step the smoothing filter removes small structures from the input image, but it also blurs strong edges. The second step tries to restore those smoothed strong edges iteratively by applying the guided edge-aware filter.\\
In Section \ref{sec:related_work} of this work we present a short review of recent edge-preserving filters which are most related to our work. Section \ref{sec:smooth_and_restore} describes our edge-preserving smoothing model in detail and gives some possible choices for the smoothing and the edge-aware filters. It further introduces a simple edge-aware filter, by using an approach similar to \cite{pham2005separable} and discusses its properties. The results of our experiments and possible applications of this new framework are presented in Section \ref{sec:results} and finally Section \ref{sec:conclusion} gives a short summary and presents possible prospective work on the topics discussed.

\section{Related work}
\label{sec:related_work}
In the past, \textit{edge-aware} filters were a popular approach to edge-preserving smoothing. These are filters that detect large intensity differences between neighbouring pixels and classify such differences as edges to be preserved (\textit{strong edges}). Low differences (\textit{weak edges}) are treated as unimportant structures and smoothed away. Early approaches include for example anisotropic diffusion \cite{perona1990scale} or the symmetric nearest neighbour filter \cite{SNNoriginal}, an example for a more recent approach is the guided filter \cite{guidedfilter}. Such filters normally ignore the size of the structures in an image, as small structures can contain strong edges as well. However, such small structures are often unwanted details that we want to smooth away. A filter that takes into account the size of an edge is called a \textit{scale-aware} filter. Edge-aware filters are therefore usually not scale-aware.\\
Whereas edge-aware filters only detect edges with large intensity difference, the bilateral filter \cite{tomasi1998bilateral}, being a combination of a Gaussian in the spatial domain and a Gaussian in the range domain, smoothes away small structures and therefore detects only strong edges larger than a certain scale, depending on its parameters. While the bilateral filter generates satisfying output images, it is very slow to compute. Therefore different optimizations have been proposed to speed up the computation (e.g. \cite{pham2005separable}, \cite{paris2009fast}, \cite{adams2010fast}), but these are only approximations to the original bilateral filter, suffering from a loss in quality or needing highly optimized code running on GPU's, especially if aiming for real-time performance.
To further enhance the bilateral filter, the joint or cross bilateral filter was introduced \cite{petschnigg2004digital}, which uses a guidance image to better remove noise or unwanted structures in an image. \\
Using this joint bilateral filter and exploiting its scale-awareness, the rolling guidance filter introduced in \cite{rollingguidance} iteratively computes a guidance image to apply the joint bilateral filter (or generally any edge-preserving smoothing filter) to the input image. It is the work that is most related to ours, as it aims for scale-aware filtering by blurring away small structures as well.

\section{Smooth and iteratively Restore}
\label{sec:smooth_and_restore}
While applying only edge-aware filters leads to good results concerning strong edges, they lack at removing small structures from the image, because those small structures often contain strong edges themselves. Therefore it is necessary to remove those small structures before applying an edge-aware filter, so that the edge-aware filter can smooth the image while preserving strong edges. To achieve this, we first apply a different filter which removes small structures, but which doesn't care much about strong edges, relieving the edge-aware filter from the task to remove small structures. This can be for example a spatial domain filter such as the Gaussian filter. However applying the edge-aware filter to the blurry output of the smoothing filter won't result in any visually appealing images, it will most probably result in worse output than after applying the edge-aware filter on the original input. But we still have the original input image, which has the best edge information we can have. By extending the edge-aware filter to use the original input image as a guidance image, we obtain a guided edge-aware filter which has no problems filtering the blurry output of the smoothing filter. It chooses the strong edges from the original input image and accordingly restores the blurry output image of the smoothing filter. This whole process can be simply described as smooth and iteratively restore (SiR).

\subsection{The Algorithm}
\begin{figure}[htbp]
	\begin{minipage}[t]{0.3\textwidth}
		\centering
		\begin{tikzpicture}[spy using outlines={circle,yellow,magnification=3,size=1.7cm, connect spies}]
			\node{\fbox{\includegraphics[width = 0.5\textwidth]{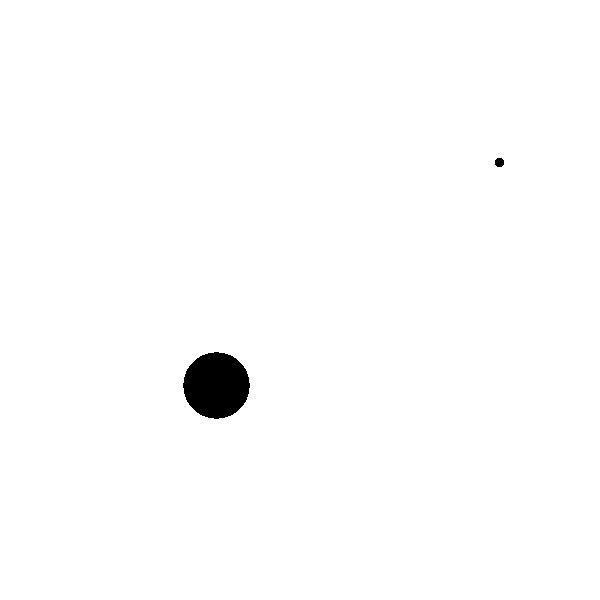}}};
			\spy on (0.793,0.548) in node [left] at (3.5,1);
			\spy on (-0.335,-0.335) in node [left] at (3.5,-1);
		\end{tikzpicture}
	\end{minipage}
	\hfill
	\begin{minipage}[t]{0.3\textwidth}
		\centering
		\begin{tikzpicture}[spy using outlines={circle,yellow,magnification=3,size=1.7cm, connect spies}]
			\node{\fbox{\includegraphics[width = 0.5\textwidth]{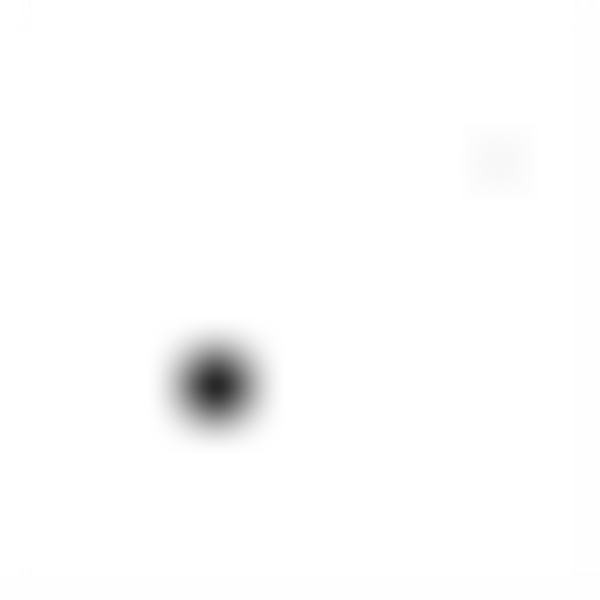}}};
			\spy on (0.793,0.548) in node [left] at (3.5,1);
			\spy on (-0.335,-0.335) in node [left] at (3.5,-1);
		\end{tikzpicture}
	\end{minipage}
	\hfill
	\begin{minipage}[t]{0.3\textwidth}
		\centering
		\begin{tikzpicture}[spy using outlines={circle,yellow,magnification=3,size=1.7cm, connect spies}]
			\node{\fbox{\includegraphics[width = 0.5\textwidth]{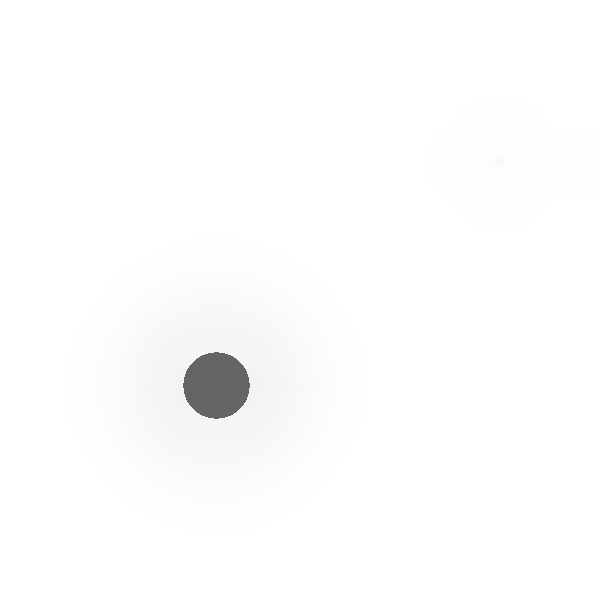}}};
			\spy on (0.793,0.548) in node [left] at (3.5,1);
			\spy on (-0.335,-0.335) in node [left] at (3.5,-1);
		\end{tikzpicture}
	\end{minipage}
	\caption{Example showing how the algorithm works. Left: original image (guidance image) with two strong edges, one is large (magnification on bottom), the other small (magnification on top). Center: blurred image, large structure still visible, small structure blurred away. Right: output of SiR after 5 iterations, large structure was restored successfully, while small structure could not be restored due to unsufficient color information in blurred image.}
	\label{pic:small_structures}
\end{figure}

In the following sections we present some possible choices for the smoothing and the guided edge-aware filter, but for now let us assume we have already chosen two such filters. Let us denote the smoothing filter with $\mathcal{S}$, the guided edge restoring filter $\mathcal{R}_{I_G}$ using a guidance image $I_G$, our input image with $I$ and our output image with $O$. Then our algorithm first performs a smoothing operation by applying $\mathcal{S}$ to the input image $I$. This smoothing operation removes small structures up to a certain scale. How much it removes depends on the parameters of $\mathcal{S}$. Any smoothing filter that removes unwanted small edges and structures can be used here, e.g. the Gaussian filter or box filter.\\
Now we have a smoothed image with already removed small structures, but the strong edges are smoothed as well. We therefore unsmooth those strong edges, while smoothed regions without strong edges are left as they are. We apply $\mathcal{R}_{I_G}$ in an iterative manner using $n$ iterations. This iterative restoring step cannot restore small edges: even though they are present in the guidance image, the necessary color information is not present any more in the blurred image. Therefore, when $\mathcal{R}_{I_G}$ tries to restore those small structures found in $I_G$, it has no color information available to do that and it will therefore not be possible to restore those edges. On the other hand, large edges are still visible in the blurred image, therefore $\mathcal{R}_{I_G}$ can still restore large edges present in $I_G$ using the color information from the blurred image (see Figure \ref{pic:small_structures} for a small example of how this works).\\
Putting those two steps together, we get the following algorithm:\\
\begin{algorithm}[H]
	\label{algo:SIR}
\KwIn{input image $I$, guidance image $I_G$, smoothing filter $\mathcal{S}$, guided edge-aware filter $\mathcal{R}_{I_G}$, number of iterations $n$}
	\KwOut{filtered image $O$}
	Blur image $I$ with smoothing filter $\mathcal{S}$: $O = \mathcal{S}(I)$\\
	\For{ $i=1$...$n$ }
	{
	filter $O$ with guided edge-aware filter $\mathcal{R}_{I_G}$: $O=\mathcal{R}_{I_G}(O)$
	}
	\caption{Smooth and iteratively restore}
\end{algorithm}
~\\
It is easily implemented as soon as you have two filters which fulfill the requirements for $\mathcal{S}$ and $\mathcal{R}_{I_G}$ and because of its very flexible nature, the algorithm can be easily adapted to the needs of your application.

\subsection{Remove small structures}
To remove small structures from an input image $I$ we apply a smoothing filter $\mathcal{S}$. This can be for example a domain filter, such as a Gaussian filter or a box filter, but can generally be any filter that smoothes away small structures. For this first step, it is not important to choose an edge-aware filter, restoring strong edges is the goal of step 2, but you can of course choose your filter to be more complex (e.g. bilateral filter). We experimented in particular with two different smoothing filters: the Gaussian filter and the box filter, which can be applied several times to approximate a Gaussian filter.

\subsection{Restore strong edges}
After removing small structures we apply an edge-aware filter to restore strong edges; this is a filter that smooths differences in pixel intensities and not in pixel coordinates. For our framework, we need to choose the filter to be guided in that sense, that it uses a guidance image to perform the filtering operation, such as the guided filter \cite{guidedfilter} or the rolling guidance filter \cite{rollingguidance}. To restore the strong edges from the input image $I$ we can just use $I$ as guidance image, but for different applications (such as flash/no-flash denoising, see \cite{petschnigg2004digital}) we might want to use other guidance images.\\
For the rest of this subsection we use following notation: the original input image (before applying $\mathcal{S}$) will be denoted $I$, the output image of the first step (this is $\mathcal{S}(I)$) will be denoted $J$ and the output image of the whole algorithm will be denoted $O$.\\
We now present three possible guided edge-aware filters to use in our framework. In the results section we show some output images generated with those filters and discuss some of their properties.

\subsubsection{2D Gaussian Range Filter}
A simple range filter, that is used for example in the bilateral filter, is the Gaussian function on the pixel intensities:
\begin{equation}
	\label{eq:simple_range}
	O(\tilde x,\tilde y) = \alpha^{-1} \sum_{(x,y)\in N(\tilde x,\tilde y)} e^{-\frac{1}{2\sigma^2} (J(\tilde x,\tilde y)-J(x,y))^2} J(x,y)
\end{equation}
with the normalization coefficient $\alpha = \sum_{(x,y)\in N(\tilde x,\tilde y)} e^{-\frac{1}{2\sigma^2} (J(\tilde x,\tilde y)-J(x,y))^2}$.
This is a weighted average working in a neighbourhood $N$ of $(\tilde x,\tilde y)$, that favours pixels $(x,y)$ with similar color. It detects strong edges (this means $(J(\tilde x,\tilde y)-J(x,y))^2$ is very high) and is therefore well suited to smooth while retaining strong edges. However to work on an already smoothed image $J$, we modify \eqnref{eq:simple_range} to use a guidance image $I_G$:
\begin{equation}
	\label{eq:guided_range}
	O(\tilde x,\tilde y) = \alpha^{-1} \sum_{(x,y)\in N(\tilde x,\tilde y)} e^{-\frac{1}{2\sigma^2} (I_G(\tilde x,\tilde y)-I_G(x,y))^2} J(x,y)
\end{equation}
with the normalization coefficient $\alpha = \sum_{(x,y)\in N(\tilde x,\tilde y)} e^{-\frac{1}{2\sigma^2} (I_G(\tilde x,\tilde y)-I_G(x,y))^2}$.
Here we still filter the image $J$, but we use $I_G$ as a guidance image. We therefore detect the strong edges of the guidance image $I_G$ and restore them in the smoothed image $J$.

\subsubsection{Recognizing pixels in the same region}
The 2D Gaussian range filter has two drawbacks: first of all, it is very slow to compute (see Table \ref{tab:runtime} for a comparison). If our goal is a fast filter for real-time applications, this filter will not suit our needs, especially if we are going to use it iteratively.\\
The second disadvantage is a bit more difficult to spot. What we actually want to achieve is some kind of segmentation: we want an edge-preserving algorithm to recognize different regions/objects in our image, to smooth intensity discrepancies inside those regions and to sharpen the borders between those regions. The Gaussian range filter detects regions only by comparing intensities of two pixels $p$ and $q$, so it actually works with following definition of a region: two pixels $p$ and $q$ are in the same region when they have nearly the same intensity value. Here $p$ and $q$ are not necessarily direct neighbours.\\ 
This is a somewhat simple definition of a region. Humans can distinguish different objects of similar color in an image as long as there is a visible border between those regions. The problem with the above definition of a region is, that two pixels $p$ and $q$ can be assigned to the same region even if they belong to different objects in the image, because this definition doesn't take into account the possible occurence of a border between $p$ and $q$.\\
To overcome this issue we introduce \textit{paths}. Given two pixels $p$ and $q$ in the image a pixel $r$ is \textit{in between} $p$ and $q$ if $d(p,r)+d(r,q) \le f\cdot d(p,q)$ for a distance measure $d(u,v)$  and a constant $f\ge 1$. \footnote{Notice that the choice $d(u,v)  = |u_x-v_x| + |u_y-v_y|$ and $f=1$ defines all pixels within the window spanned by $p$ and $q$ as between $p$ and $q$}. We say a \textit{path} from $p$ to $q$ is a finite sequence of not necessarily neighbouring pixels $t_0$, ..., $t_k$ for some $k>0$, where $t_0 := p$ and $t_k := q$, such that for every $0<i<k$, $t_i$ is in between $t_{i-1}$ and $t_{i+1}$. We call $t_1$, ..., $t_{k-1}$ the intermediate pixels between $p$ and $q$. Furthermore we say that there is an \textit{edge} between two consecutive pixels of the path $t_i$ and $t_{i+1}$ if there is an intensity difference $(I(t_i)-I(t_{i+1}))^2>0$. This edge is called a \textit{weak edge} if the difference is low and a \textit{strong edge} if there is a large difference. Finally we say that there is a strong edge along the path from $p$ to $q$ if there exists an $i\ge 0$, such that there is a strong edge between $t_i$ and $t_{i+1}$.

With these definitions we are now able to come up with a better definition for a region: two pixels $p$ and $q$ with a predefined path $t_0$, ..., $t_k$ lie in the same \textit{region} if there is no strong edge along this path.\\
Notice however that we didn't specify how exactly this path should be chosen. Different paths from $p$ to $q$ may lead to different regions, so choosing a reasonable path is crucial for defining a useful region.\\
Our first definition of a region is a special case of the second definition: for all pixels $p$ and $q$ it simply considers the trivial path consisting of only $p$ and $q$. To see the difference compared to the general case with a path $t_0$, ..., $t_k$, $k>1$, we can distinguish four cases:
\begin{enumerate}[I]
	\item All $t_i$ have almost the same intensity, so both $p$ and $q$ lie in the same region according to both definitions. \label{enum:1}
	\item $p$ and $q$ have different values and there exists an $i$ such that $0\le i< k$ where the intensity difference $(I(t_i)-I(t_{i+1}))^2$ is very high. According to both definitions $p$ and $q$ lie in different regions. \label{enum:2}
	\item If the path from $p$ to $q$ is a monotonic ramp, i.e. if $I(p)<I(t_1)<...<I(t_{k-1})<I(q)$ or $I(p)>I(t_1)>...>I(t_{k-1})>I(q)$ such that there is a high intensity difference $(I(p)-I(q))^2$ between $p$ and $q$ but low intensity differences $(I(t_i)-I(t_{i+1}))^2$ between consecutive points of the path, then the first definition would place $p$ and $q$ in different regions, whereas the second definition would recognize no strong edge but rather a smooth transition from $p$ to $q$ and would therefore place them in the same region. \label{enum:3}
	\item Finally if $p$ and $q$ have similar values but there exists an $i$ such that $0\le i< k$ where $(I(t_i)-I(t_{i+1}))^2$ is very high, then we have at least one strong edge along $t_0$, ..., $t_k$. According to the first definition of a region, $p$ and $q$ belong to the same region but the second definition places them in two different regions. \label{enum:4}
\end{enumerate}
Only in the first two cases the first definition of a region leads to a desired classification. In the other two cases it doesn't, because it treats \ref{enum:3} as \ref{enum:2} and \ref{enum:4} as if it were \ref{enum:1}.\\
Using paths, we can better classify the relationship between $p$ and $q$. The next filter uses a simple path with $d(u,v)  = |u_x-v_x| + |u_y-v_y|$, $f=1$ and one intermediate pixel $t_1$.

\subsubsection{Separable Gaussian Range Filter}
\begin{figure}[htbp]
	\centering
	\begin{minipage}[t]{0.4\textwidth}
		\begin{minipage}[t]{0.49\textwidth}
			\centering
			\includegraphics[width = 0.5\textwidth]{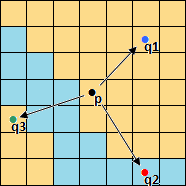}
			\label{pic:2Dgauss}
		\end{minipage}
		\hfill
		\begin{minipage}[t]{0.49\textwidth}
			\centering
			\includegraphics[width = 0.5\textwidth]{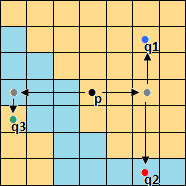}
		\end{minipage}
		\caption{Schematic view of the 2D Gaussian range filter (left) and the separable Gaussian range filter (right).}
		\label{pic:separable}
	\end{minipage}
	\hfill
	\begin{minipage}[t]{0.2\textwidth}
		\centering
		\includegraphics[width = 0.5\textwidth]{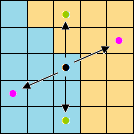}
		\label{pic:SNN}
		\caption{Schematic view of the SNN filter with a $5\times 5$ neighbourhood.}
	\end{minipage}
	\hfill
	\begin{minipage}[t]{0.2\textwidth}
		\centering
		\includegraphics[width = 0.5\textwidth]{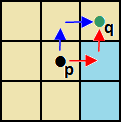}
		\caption{Example showing that $\mathcal{O}_{h,I}$ and $\mathcal{O}_{v,I}$ are not commutative.}
		\label{pic:notcommutative}
	\end{minipage}
\end{figure}

We now proceed similarly to \cite{pham2005separable}, who approximate the 2D bilateral filter by simply applying a 1D bilateral filter first horizontally and then vertically. But instead of approximating the whole bilateral filter, we only approximate the Gaussian range filter \eqnref{eq:guided_range}.
First we define two operators $\mathcal{O}_{h,I}$ and $\mathcal{O}_{v,I}$ as follows:
\begin{align}
	\mathcal{O}_{h,I}(J)(\tilde x,y) &= \alpha_1^{-1} \sum_{x\in N_1(\tilde x)} e^{-\frac{1}{2\sigma^2} (I(\tilde x,y)-I(x,y))^2} J(x,y)\\
	\mathcal{O}_{v,I}(J)(x,\tilde y) &= \alpha_2^{-1} \sum_{y\in N_2(\tilde y)} e^{-\frac{1}{2\sigma^2} (I(x,\tilde y)-I(x,y))^2} J(x,y)
\end{align}
with normalization coefficients $\alpha_1$ and $\alpha_2$ and neighbourhoods $N_1$ of $\tilde x$ and $N_2$ of $\tilde y$.
Those two operators compute a Gaussian over the pixel intensities, but they do it only in one dimension: $\mathcal{O}_{h,I}$ filters horizontally and $\mathcal{O}_{v_I}$ vertically. 
The combination $\mathcal{O}_{h,I}(\mathcal{O}_{v,I}(J))$ of these two operators leads to a filter that can 
be written in a closed form as
\begin{equation}
	\begin{aligned}
		O(\tilde x,\tilde y) &= \alpha^{-1} \sum_{(x,y) \in N(\tilde x,\tilde y)}  e^{-\frac{1}{2\sigma^2} \big(\left(I(\tilde x,\tilde y)-I(x,\tilde y)\right)^2 + (I(x,\tilde y)-I(x,y))^2\big) } J(x,y)
	\end{aligned}
\end{equation}
with $\alpha$ being the appropriate normalization coefficient and a neighbourhood $N$ of $(\tilde x,\tilde y)$. There are two important aspects to note here:
\begin{itemize}
		\label{item:lab}
	\item The resulting filter is by definition separable and therefore fast to compute.
	\item We could also try $\mathcal{O}_{v,I}(\mathcal{O}_{h,I}(J))$ which would result in a new filter because $\mathcal{O}_{h,I}$ and $\mathcal{O}_{v,I}$ are not commutative. An example can be seen in Figure \ref{pic:notcommutative}: all neighbouring pixels of $p$ get the same weight for both paths, except for $q$ which gets a different weight. The filter $\mathcal{O}_{h,I}(\mathcal{O}_{v,I})$ does not detect a boundary between $p$ and $q$ whereas $\mathcal{O}_{v,I}(\mathcal{O}_{h,I})$ uses a different path which has a boundary.
\end{itemize}
In contrast to the 2D Gaussian range filter, this filter doesn't directly compare $I(x,y)$ and $I(\tilde x,\tilde y)$ but rather uses an intermediate pixel $I(x,\tilde y)$. Using the notation from above we have $t_0 = p=(\tilde x,\tilde y)$, $t_1 = (x,\tilde y)$ and $t_2 = q=(x,y)$. The distance measure and constant that are necessary to define the path that we use here are $d(u,v)  = |u_x-v_x| + |u_y-v_y|$ and $f=1$. \\
To see how this separable filter differs from the 2D Gaussian range filter, we look at the computed weights in these two filters:
\begin{align}
	w_{1,(p,q)} &= e^{-\frac{1}{2\sigma^2} (I(p)-I(q))^2}\\
	w_{2,(p,q)}	&=e^{-\frac{1}{2\sigma^2}\big((I(p)-I(t_1))^2 + (I(t_1)-I(q))^2\big) } = w_{1,(p,q)}\cdot e^{-\frac{1}{\sigma^2}(I(t_1)-I(q))\cdot (I(t_1)-I(p)) }
	\label{eq:w_new}
\end{align}
Analogous to \ref{enum:1} - \ref{enum:4} this leads us to three different cases:
\begin{itemize}
	\item $I(p)=I(t_1)\vee I(q)=I(t_1)$ $\Rightarrow~w_{1,(p,q)}=w_{2,(p,q)}$:  ~If the intermediate pixel $t_1$ equals $p$ or $q$, then the resulting weight is the same as in the first approach. This corresponds to the cases \ref{enum:1} and \ref{enum:2}.
	\item $I(p)<I(t_1)<I(q)\vee I(p)>I(t_1)>I(q)$ $\Rightarrow~w_{1,(p,q)}>w_{2,(p,q)}$: ~If the intermediate pixel $t_1$ has an intensity value between $p$ and $q$ then although there may be a (large) difference between $p$ and $q$ (which the 2D Gaussian range filter would classify as an edge), this filter increases the resulting weight. That's because although they differ, the intermediate pixel tells us they belong to the same region and that there's a smooth transition from $p$ to $q$. This corresponds to \ref{enum:3}.
	\item $\big(I(t_1)>\max{\{I(p),I(q)\}}\big)\vee \big(I(t_1)<\min{\{I(p),I(q)\}}\big)$ $\Rightarrow~w_{1,(p,q)}<w_{2,(p,q)}$: ~If the intermediate pixel $t_1$ has a larger or smaller intensity value than both $p$ and $q$, then there is an edge between $p$ and $q$ and the resulting weight is therefore smaller. This corresponds to \ref{enum:4}.
\end{itemize}
The path $t_0,t_1,t_2$ that we chose here leads to a reasonable description of a region for small neighbourhoods $N(\tilde x, \tilde y)$. We therefore get a filter that is not only separable and therefore faster to compute than the 2D Gaussian range filter, but that also uses an intermediate pixel to check if two pixels $p$ and $q$ lie in the same region according to a path (see Figure \ref{pic:separable} for an illustration of how the separable Gaussian filter works. Note how the 2D Gaussian range filter doesn't recognize the edge between $p$ and $q3$ whereas the separable Gaussian range filter correctly recognizes it.).

\subsubsection{Symmetric Nearest Neighbour Filter}
The third edge-aware filter we used is the symmetric nearest neighbour filter (SNN) first presented in \cite{SNNoriginal}. It compares all pixels in a predefined neighbourhood of a pixel $p$ and chooses of each pixel pair of opposite pixels the more similar neighbour to $p$. $p$ is then replaced by the mean (SNNmean) or the median (SNNmedian) of the symmetric pixels more similar to the center. In the simplest case we consider a $3\times 3$ neighbourhood, therefore it compares the eight neighbouring pixels of $p$ and calculates the mean or median of four pixels. This leads to a very fast edge-aware filter. In our results section we considered only this simple $3\times 3$ case.

\subsection{Advantages and Disadvantages}
\label{sec:disadvantage}
The main advantage of this framework is that it is very flexible. According to the application field one can choose which filters are most suitable. Removing noise from an image may require a different guided filter than texture extraction or pre-processing for edge detection.\\
Furthermore the SiR filter is very easy to implement. If you already have a smoothing and an edge-aware filter, then the implementation is achieved with only a few lines of code. The performance depends directly on the performance of both filters. Plugging an edge-aware filter into SiR will necessarily take more time than applying only the edge-aware filter alone, but it may improve the overall performance.\\
There are however two drawbacks using this framework: First of all, as blurring may lead to less intense colors, this unwanted property transfers to the output image too, as the output image is a restoration based on the blurred image. While the resulting image consists of smooth regions and sharp edges, it may also lose some of its color intensity, especially if the whole algorithm is applied more than only one time. The second disadvantage concerns the restoration of small edges near large edges. Near large edges, which still exist after blurring, there exists color information to restore not only the large edge itself, but also surrounding structures with similar color, it is possible that small edges reappear near large edges after blurring them away. This means that near large structures, small structures or small artifacts can appear, because the edge-aware filter can partially restore them due to the available color information from the large edge.

\subsection{Comparison to the Rolling Guidance Filter}
\label{sec:rgf}
RGF and SiR share the same main idea: remove small structures with an appropriate filter and use an edge-aware filter to handle strong edges. However they differ in how they implement this idea. The RGF fixes the input image, each iteration is filtering the same image. In every iteration it computes a new guidance image to better filter the input image in the next iteration, starting with a blurred input image in the first step. The SiR on the other hand does the reverse: it fixes the guidance image, using its optimal edge information, and instead modifies the image to be filtered in every iteration. It starts off with an initially blurred image and restores the strong edges according to the original input image. This differing setting of which image is fixed and which one changes iteratively leads to two different principles: calculate a reasonable guidance image to filter the input image once (RGF) versus use all the edge information available as guidance to restore important edges in a blurred image (SiR). 
In Section \ref{sec:results} you can see comparisons concerning the output of both algorithms.

\section{Applications and Results}
\label{sec:results}
In this section we present results obtained with our framework during different experiments and show some possible applications. When calculating the Gaussian blur, the 2D Gaussian range filter or our separable Gaussian range filter, we always used a $7\times7$ neighbourhood. For calculating the SNN we used a $3\times 3$ neighbourhood. We applied our algorithm to each color channel separately in $RGB$ color space.\\
All results in this section where obtained on an Intel $i7$ processor with 2.9 GHz. Our C++ implementation uses no SIMD instructions and all implementations were running on a single thread, therefore there is still a lot of room for improvement. However, even this simple implementation results in a very fast filtering operation. The results are compared to the C++ implementation of the rolling guidance filter \cite{rollingguidance} using the joint bilateral filter which is freely available from their project homepage. It uses only one single thread and no SIMD instructions as well. However to speed up the bilateral filtering it implements the permutohedral lattice from \cite{adams2010fast}.

\subsection{Runtime evaluation}
\label{sec:runtime}
Before we look at the actual output of our algorithm, we compare the runtimes of the rolling guidance filter (RGF) and different filter combinations using our model (SiR). In Table \ref{tab:runtime} we applied all listed filters to the image shown in Figure \ref{pic:mosaic_comparison}. For RGF there is only a total runtime entry as it doesn't separate the smooth and restore steps. The parameters were chosen as follows:
\begin{itemize}
		\item RGF: $\sigma_s=3$, $\sigma_r=25$, $4$ iterations
		\item SiRSNN: SiR with iterated Box blur ($5\times 5$ neighbourhood, applied 2 times) and SNNmean.
		\item SiR2DGauss: SiR with Gaussian blur ($\sigma = 5$) and 2D Gaussian filter ($\sigma=20$)
		\item SiRsep: SiR with Gaussian blur ($\sigma=5$) and separable Gaussian filter ($\sigma=20$)
\end{itemize}
For SiR2DGauss and SiRsep we set the numbers of iterations to $n=5$, for SiRSNN we set $n=9$.\\
As we can see, SiR yields far better runtime results than the rolling guidance filter, the box blur and SNN combination even results in 10fps real-time performance without any kind of optimization. Even our slowest method, SiR2DGauss, runs in only a third of the time of the RGF.
\begin{table}[htp]
	\centering
	\begin{tabular}{|c|c|c|c|c|}
		\hline
		   & RGF & SiR2DGauss & SiRsep & SiRSNN\\ \hline
		Smooth step & X & $0.05s$ & $0.05s$ & $0.03s$\\ \hline
		Iteratively Restore step & X & $0.72s$ & $0.29s$& $0.07s$\\ \hline
		Total & $2.4s$ & $0.77s$ & $0.34s$ & $0.1s$\\ \hline
	\end{tabular}
	\caption{Runtime comparison on the image seen in Figure \ref{pic:mosaic_comparison} with dimensions 802$\times$682 using no optimizations.}
	\label{tab:runtime}
\end{table}

\subsection{Texture smoothing}
\begin{figure}[th]
	\centering
	\begin{minipage}[t]{0.99\textwidth}
		\centering
		\begin{tikzpicture}[spy using outlines={magnification=5,size=2.5cm}]
			\node{\includegraphics[width = 0.17\textwidth]{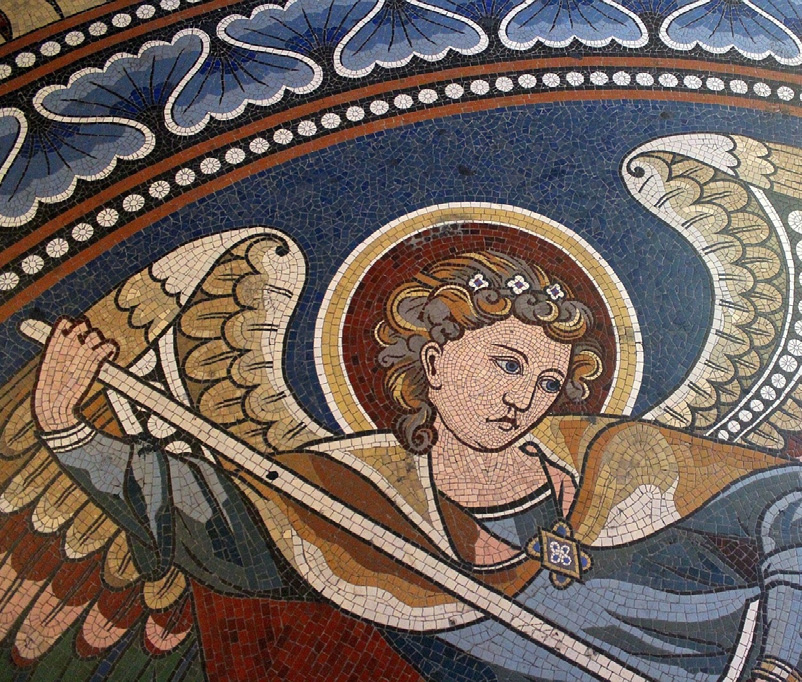}};
			\spy [yellow] on (0.793,0.548) in node [left] at (1.3,-2.6);
			\spy [blue] on (0.3,-0.1) in node [left] at (1.3,-5.4);
		\end{tikzpicture}
		\begin{tikzpicture}[spy using outlines={magnification=5,size=2.5cm}]
			\node{\includegraphics[width = 0.17\textwidth]{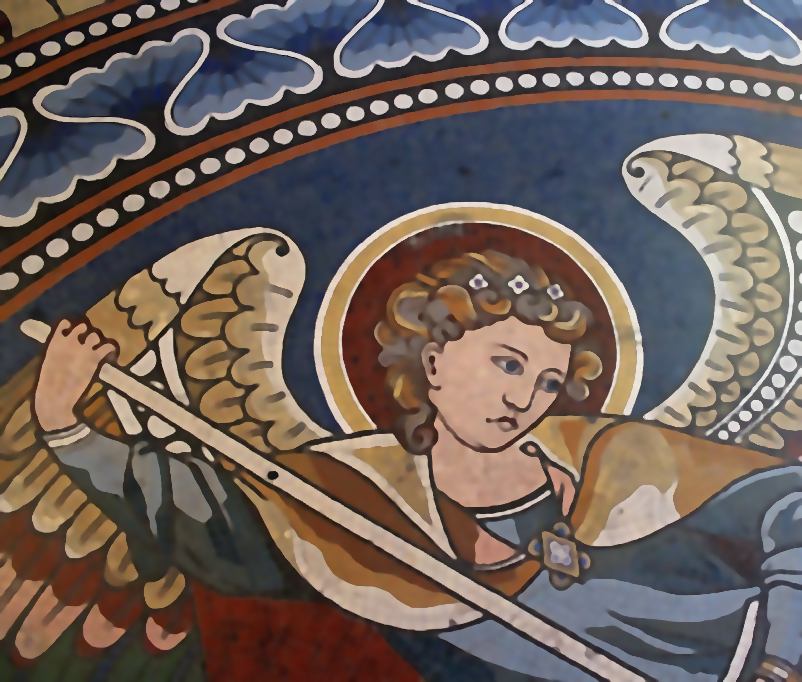}};
			\spy [yellow] on (0.793,0.548) in node [left] at (1.3,-2.6);
			\spy [blue] on (0.3,-0.1) in node [left] at (1.3,-5.4);
		\end{tikzpicture}
		\begin{tikzpicture}[spy using outlines={magnification=5,size=2.5cm}]
			\node{\includegraphics[width = 0.17\textwidth]{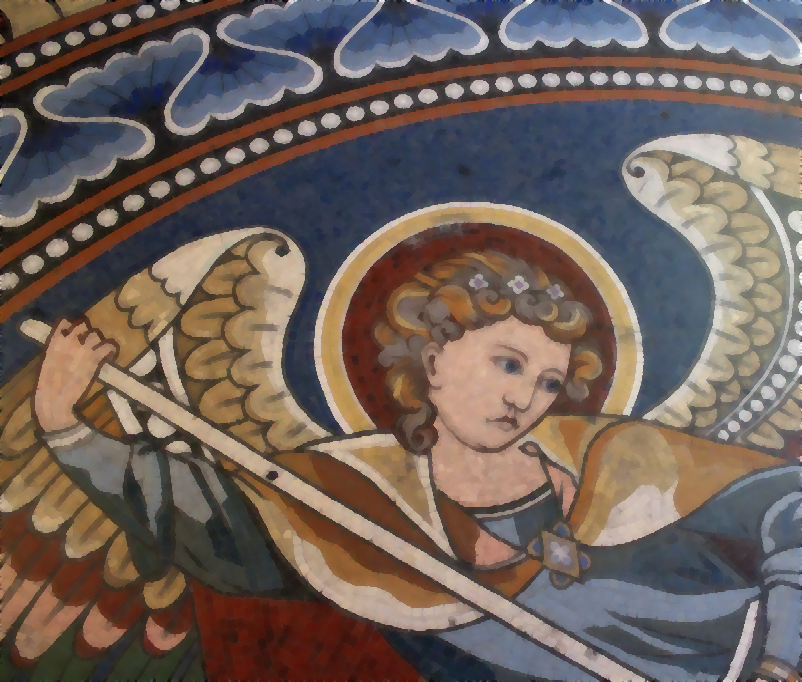}};
			\spy [yellow] on (0.793,0.548) in node [left] at (1.3,-2.6);
			\spy [blue] on (0.3,-0.1) in node [left] at (1.3,-5.4);
		\end{tikzpicture}
		\begin{tikzpicture}[spy using outlines={magnification=5,size=2.5cm}]
			\node{\includegraphics[width = 0.17\textwidth]{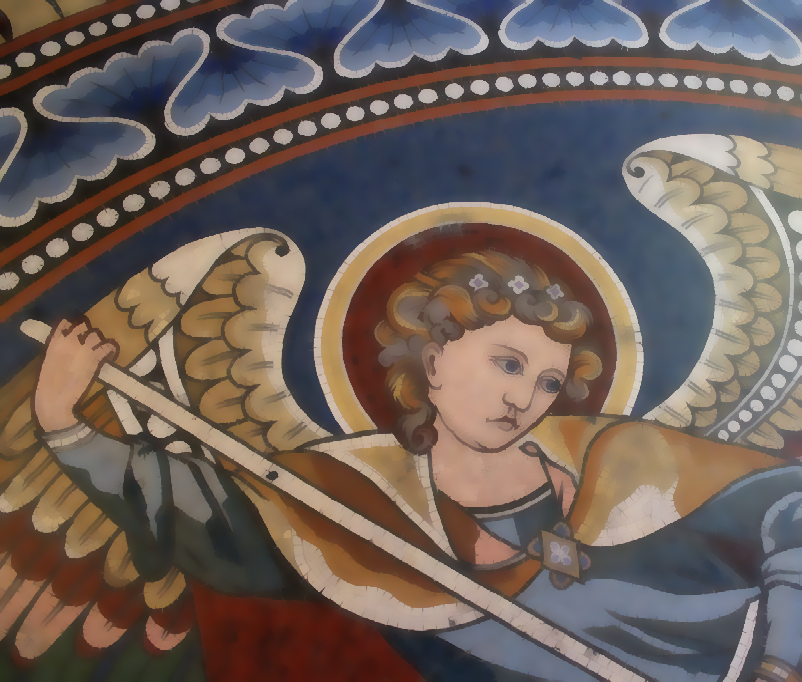}};
			\spy [yellow] on (0.793,0.548) in node [left] at (1.3,-2.6);
			\spy [blue] on (0.3,-0.1) in node [left] at (1.3,-5.4);
		\end{tikzpicture}
		\begin{tikzpicture}[spy using outlines={magnification=5,size=2.5cm}]
			\node{\includegraphics[width = 0.17\textwidth]{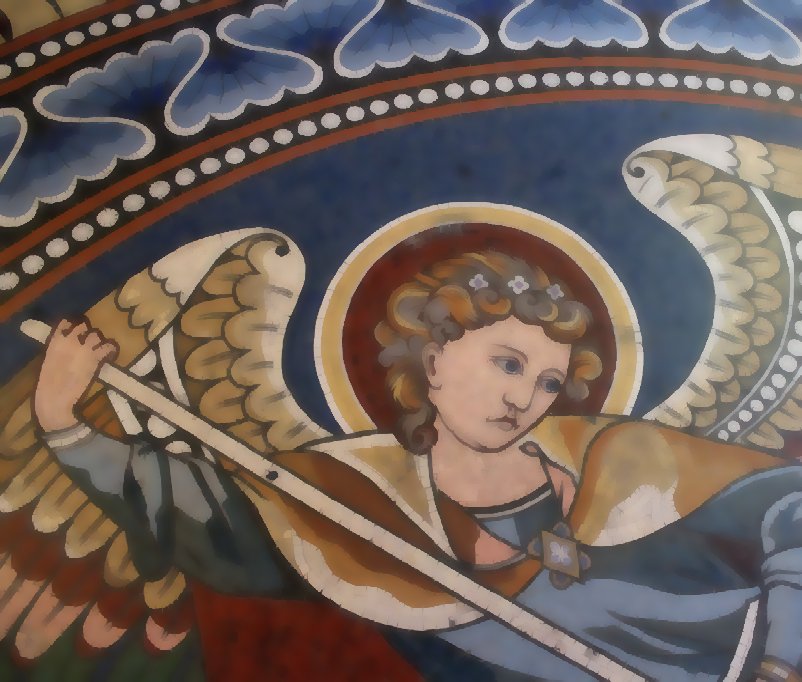}};
			\spy [yellow] on (0.793,0.548) in node [left] at (1.3,-2.6);
			\spy [blue] on (0.3,-0.1) in node [left] at (1.3,-5.4);
		\end{tikzpicture}
		\caption{Texture smoothing output of different filters, from left to right: original image, RGF, SiRSNN, SirGauss2D, SiRsep. Parameters are listed in Section \ref{sec:runtime}. The image was taken from \cite{rollingguidance}.}
		\label{pic:mosaic_comparison}
	\end{minipage}
\end{figure}

The smooth-and-iteratively-restore (SiR) framework can be used for texture smoothing or texture extraction. We tested it on various images and compared it to the rolling guidance filter (RGF), which is the only state-of-the-art scale-aware edge-preserving smoothing filter that we know of. 
All obtained results where very similar to those shown in Figure \ref{pic:mosaic_comparison}:
In terms of smoothing regions with similar color, SiRsep achieves more homogeneous results than RGF and SiRSNN, which both produce bright stains. 
Furthermore the overall appearance of the edges with SiRsep seems better in terms of preserving the original shape of the edges. RGF produces smoother but also less detailed edges, while SiR tries generally to restore the edges as similar as possible to the original image.\\
However, as mentioned in Section \ref{sec:disadvantage}, in the neighbourhood of edges SiRsep produces some artifacts because it partially restores small structures, while RGF doesn't. Also, the resulting image of SiR suffers from less color intensity than the original one because of the smoothing step.\\
SiRSNN produces noisier results than both SiRsep and RGF, however in terms of speed it is way ahead. If aiming for real time applications, SiRSNN may still satisfy your requirements while being extremely fast.\\
If the output of RGF is more desirable, but the performance is too slow, one can combine our method with RGF: use SiR to compute a smoothed image with very good edge information fast and then use that image as initial guidance image in RGF and perform therefore only a few steps (ideally only one) with RGF.

\subsection{Pre-processing for edge detection}
\begin{figure}[htp]
	\centering
	\begin{minipage}[t]{0.99\textwidth}
		\centering
		\includegraphics[width = 0.24\textwidth]{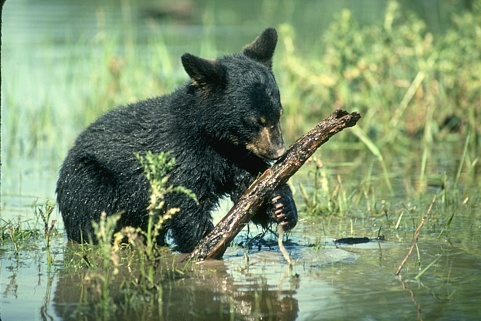}
		\includegraphics[width = 0.24\textwidth]{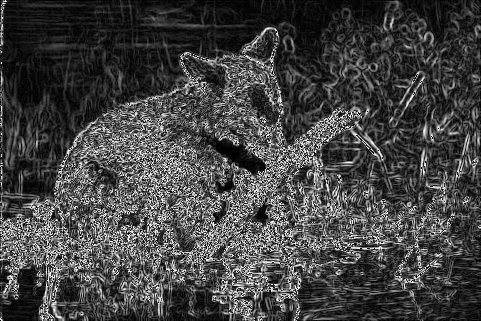}
		\includegraphics[width = 0.24\textwidth]{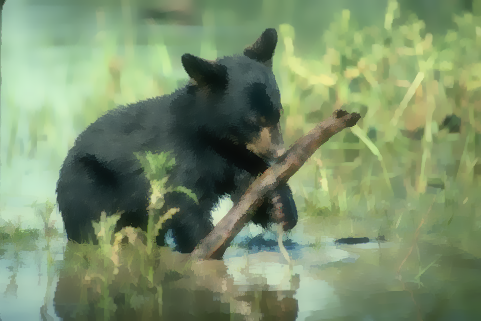}
		\includegraphics[width = 0.24\textwidth]{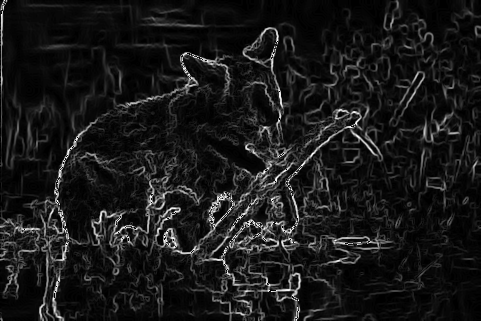}
	\end{minipage}
	\caption{Sample results of edge detection before and after applying SiRsep. Image taken from \cite{MartinFTM01}.}
	\label{pic:edgedetect}
\end{figure}

Edge detection is a good example where small structures with strong edges can mislead the used edge detection algorithm. In such cases using a scale-aware edge-preserving filter can improve the result drastically by removing those small structures while retaining the important edges (for an example see Figure \ref{pic:edgedetect}).
We tested our framework on the \textit{Berkeley Segmentation Dataset} \cite{MartinFTM01} together with the \textit{Sobel} edge detector. We tested 3 different settings: edge detection without pre-processing, edge detection after using one of the three edge-aware filters and finally edge detection using those same edge-aware filters plugged in in our framework. As can be seen in Table \ref{tab:bsds_sobel} our algorithm does not only achieve significant better results than edge detection without pre-processing, it also achieves clearly better results than using one of the edge-aware filters only as pre-processing without the added scale-awareness provided by our method. For the edge-aware filters we used following parameters:
\begin{itemize}
	\item SNN: $3\times 3$ neighbourhood
	\item 2D Gaussian Range Filter: $7\times 7$ neighbourhood, $\sigma = 8$
	\item Separable Gaussian Range Filter: $7\times 7$ neighbourhood, $\sigma = 8$
\end{itemize}
When plugged into our framework we used the same parameters as above and additionally in each case we used a Gaussian blur in a $7\times 7$ neighbourhood with $\sigma = 3$ using 5 iterations.

\begin{table}[htp!]
	\centering
	\begin{tabular}{|c|c|c|c|c|}
		\hline
		    & No filter &  SNN & 2D-Gauss & sep. Gauss\\ \hline
		Without SiR & 0.49 & 0.53 & 0.51 & 0.51\\ \hline
		With SiR & X & 0.56 & 0.56 & 0.57\\ \hline
	\end{tabular}
	\caption{Benchmark results of \textit{Sobel} edge detection using no pre-processing vs. edge-aware filter as pre-processing step vs. our method as pre-processing step on the \textit{Berkeley Segmentation Dataset}. The computed score represents the F-measure.}
	\label{tab:bsds_sobel}
\end{table}

\section{Conclusion and further work}
\label{sec:conclusion}
In this paper we presented a novel framework for edge-preserving smoothing, which uses a smoothing filter for small structure removal and a guided edge-aware filter to restore strong edges. It can be used with a variety of different filters, of which a small subset was presented within this paper. The algorithm is very simple and therefore easy to implement. It consists of two steps: First smooth once using the smoothing filter, then iteratively apply the guided edge-aware filter. Depending on the performance of the two filters and the predefined number of iterations, the algorithm can be very fast. As our experiments show, with carefully chosen filters our framework is well suited for real-time applications. \\
Further work on this topic includes experimenting with more guided edge-aware filters and applying the model to other applications. 

\bibliography{literature}{}

\begin{thebibliography}{10}

\bibitem{adams2010fast}
Andrew Adams, Jongmin Baek, and Myers~Abraham Davis.
\newblock Fast high-dimensional filtering using the permutohedral lattice.
\newblock In {\em Computer Graphics Forum}, volume~29, pages 753--762. Wiley
  Online Library, 2010.

\bibitem{SNNoriginal}
David Harwood, Muralidhara Subbarao, Hannu Hakalahti, and Larry~S Davis.
\newblock A new class of edge-preserving smoothing filters.
\newblock {\em Pattern Recognition Letters}, 6(3):155--162, 1987.

\bibitem{guidedfilter}
Kaiming He, Jian Sun, and Xiaoou Tang.
\newblock Guided image filtering.
\newblock In {\em Computer Vision--ECCV 2010}, pages 1--14. Springer, 2010.

\bibitem{MartinFTM01}
D.~Martin, C.~Fowlkes, D.~Tal, and J.~Malik.
\newblock A database of human segmented natural images and its application to
  evaluating segmentation algorithms and measuring ecological statistics.
\newblock In {\em Proc. 8th Int'l Conf. Computer Vision}, volume~2, pages
  416--423, July 2001.

\bibitem{paris2009fast}
Sylvain Paris and Fr{\'e}do Durand.
\newblock A fast approximation of the bilateral filter using a signal
  processing approach.
\newblock {\em International Journal of Computer Vision}, 81(1):24--52, 2009.

\bibitem{perona1990scale}
Pietro Perona and Jitendra Malik.
\newblock Scale-space and edge detection using anisotropic diffusion.
\newblock {\em Pattern Analysis and Machine Intelligence, IEEE Transactions
  on}, 12(7):629--639, 1990.

\bibitem{petschnigg2004digital}
Georg Petschnigg, Richard Szeliski, Maneesh Agrawala, Michael Cohen, Hugues
  Hoppe, and Kentaro Toyama.
\newblock Digital photography with flash and no-flash image pairs.
\newblock {\em ACM transactions on graphics (TOG)}, 23(3):664--672, 2004.

\bibitem{pham2005separable}
Tuan~Q Pham and Lucas~J Van~Vliet.
\newblock Separable bilateral filtering for fast video preprocessing.
\newblock In {\em Multimedia and Expo, 2005. ICME 2005. IEEE International
  Conference on}, pages 4--pp. IEEE, 2005.

\bibitem{tomasi1998bilateral}
Carlo Tomasi and Roberto Manduchi.
\newblock Bilateral filtering for gray and color images.
\newblock In {\em Computer Vision, 1998. Sixth International Conference on},
  pages 839--846. IEEE, 1998.

\bibitem{rollingguidance}
Qi~Zhang, Xiaoyong Shen, Li~Xu, and Jiaya Jia.
\newblock Rolling guidance filter.
\newblock In {\em Computer Vision--ECCV 2014}, pages 815--830. Springer, 2014.

\end{thebibliography}
\end{document}